# A Novel Driver Distraction Behavior Detection Method Based on Self-supervised Learning with Masked Image Modeling

Yingzhi Zhang, *student member*, *IEEE*, Taiguo Li, Chao Li, and Xinghong Zhou.

*Abstract*— Driver distraction causes a significant number of traffic accidents every year, resulting in economic losses and casualties. Currently, the level of automation in commercial vehicles is far from completely unmanned, and drivers still play an important role in operating and controlling the vehicle. Therefore, driver distraction behavior detection is crucial for road safety. At present, driver distraction detection primarily relies on traditional convolutional neural networks (CNN) and supervised learning methods. However, there are still challenges such as the high cost of labeled datasets, limited ability to capture high-level semantic information, and weak generalization performance. In order to solve these problems, this paper proposes a new self-supervised learning method based on masked image modeling for driver distraction behavior detection. Firstly, a self-supervised learning framework for masked image modeling (MIM) is introduced to solve the serious human and material consumption issues caused by dataset labeling. Secondly, the Swin Transformer is employed as an encoder. Performance is enhanced by reconfiguring the Swin Transformer block and adjusting the distribution of the number of window multi-head self-attention (W-MSA) and shifted window multi-head self-attention (SW-MSA) detection heads across all stages, which leads to model more lightening. Finally, various data augmentation strategies are used along with the best random masking strategy to strengthen the model's recognition and generalization ability. Test results on a large-scale driver distraction behavior dataset show that the self-supervised learning method proposed in this paper achieves an accuracy of 99.60%, approximating the excellent performance of advanced supervised learning methods. Our code is publicly available at github.com/RockyIsalady-killer/SL-DDBD.

*Index Terms*—Self-supervised learning, vision transformer, masked image modeling, driver distraction.

## I. INTRODUCTION

Driver behavior and decision control of vehicles are the main factors affecting safe driving. According to a report published by the World Health Organization (WHO), approximately 1.35 million people worldwide are killed in traffic accidents each year. Between 20 and 50 million people are injured and become disabled because of traffic accidents [1]. Studies also indicate that driver distractions are one of the important causes of road traffic accidents. Additionally, according to data from the National Highway Traffic Safety Administration (NHTSA) [2] in the United States, driver distractions contributed to 3,142 traffic accidents in 2019. The American Society of Automotive Engineers (SAE) divides autonomous driving into six levels, ranging from L0 to L5. By 2030, the United States, Europe, and China will have 80 million L4/L5 intelligent vehicles. Although autonomous driving technology has made impressive progress, the control of vehicles by autonomous driving is still immature [3]. In the autonomous driving tests conducted by Uber, 37 traffic accidents were related to driver distraction. The driver's distracting behavior caused the vehicle to not take control in emergency situations and implement emergency remedial actions in a timely manner [4]. Therefore, whether it is autonomous driving or manual driving, the driver needs to remain focused during the vehicle's journey. The driver's state is particularly important, and an efficient and accurate driver distraction detection system is an important research method for achieving traffic safety [5]. Driver distraction detection will be integrated into Advanced Driver Assistance Systems (ADAS), analyzing the driver's actions and behavior to predict unsafe distraction operations [6]. When the driver's distraction is detected, the vehicle dashboard displays a prompt message, emits a sound or lowers the windows to alert the driver. The vehicle control priority can be temporarily adjusted to initiate braking and avoid risks [7]. In previous surveys, with the help of such precise driver distraction detection systems, the likelihood of vehicle accidents on the road can be reduced by 10% to 20% [8].

The NHTSA defines driver distraction as "any activity that diverts a driver's attention away from the primary task of driving." The Centers for Disease Control and Prevention (CDC) also defines distracted driving more broadly [9]. When a driver's attention is diverted from the driving task to another activity, their attention is considered to be distracted. Cognitive distraction, behavioral distraction, and visual distraction are three typical types of driver distraction [10]. The driver is the

This research is supported by the science and technology program of Gansu Province under Grant No. 21JR7RA303, in part by Gansu Provincial Department of Education: University Teacher Innovation Fund Project under Grant No. 2023A-039, in part by Lanzhou Jiaotong University Youth Science Foundation Project under Grant No. 2020002. (Corresponding author: Taiguo Li.)

Yingzhi Zhang, Taiguo Li, and Xinghong Zhou are with College of Automation and Electrical Engineering, Lanzhou Jiaotong University, Lanzhou 730070, China (email: 12211496@stu.lzjtu.edu.cn; leetg@mail.lzjtu.edu.cn; 12221581@stu.lzjtu.edu.cn).
Chao Li is with the Shaanxi Kangfu Hospital, Xian 710065, China (email: 544738669@qq.com).

primary decision-maker and controller of the vehicle [11], so behavior distractions can greatly impact normal driving and easily result in traffic accidents. In recent years, the development of In-Vehicle Information System (IVIS) has made the infotainment and communication functions in vehicles smoother and visually rich, further attracting the driver's attention [12]. On the other hand, daily life also requires people to have a great need for communication with their phones, and some drivers even need to frequently answer calls and send texts while driving [13]. These factors have also caused the driver's focus to become more and more dispersed.

Among the methods for driver distraction detection, researchers commonly use methods such as physiological signals [14] [15] [16], vehicle information [17] [18] [19], and computer vision [20] [21] [22]. Most physiological signal detection methods have high hardware costs, and invasive physiological sensing sensors often affect the driving experience of drivers [23]. Methods based on vehicle information also have certain limitations. Hardware failures of vehicles and sensors, as well as external interference, can result in inaccurate information collection, affecting the accuracy of the system [24]. These methods also require extensive data processing and advanced algorithm support [25]. To address these issues, researchers have gradually shifted their focus to computer vision methods [26]. Computer vision-based techniques for detecting driver distraction are non-intrusive, thus they do not affect the normal driving of drivers, and they are not as inaccurate as vehicle information methods due to information errors [27]. However, most researchers primarily adopt supervised learning methods and traditional CNN models. By surveying the research work in the field of driver distraction behavior detection, three existing challenges can be inferred:

(a) The strong dependence on labeled datasets for driver distraction behavior detection based on supervised learning is a major drawback. Supervised learning models require a large amount of labeled data for training, which significantly increases the cost of training. Additionally, labeling the dataset requires a significant amount of manpower and resources. Furthermore, the complexity of real-world driving scenarios makes it difficult to accurately label data, increasing the difficulty and cost of creating a labeled distraction driving dataset.

(b) The traditional CNN-based driver distraction behavior detection model is inefficient in feature extraction and has limited ability to capture overall image information. CNN models only stack convolution and pooling operations. As a result, it is inefficient in capturing and representing long-term dependencies in the data. Additionally, CNN tend to focus on low-level image features, while the task of detecting driver behavior requires better control of global information. In images of driver distractions, basic visual pixels are correlated into objects, and the spatial relationships between objects form scene information. In such a visual detection task that requires behavior judgment, mastering high-level visual semantic information is more important.

(c) Supervised learning models based on CNN have limited capability to learn representations and show an average generalization ability. Moreover, they are not effective at focusing attention on crucial regions to detect distractions. Furthermore, their dependence on labeled datasets decreases their ability to generalize and transfer to novel tasks. This makes it challenging to ensure accurate recognition in more intricate and diverse driving scenarios.

Due to the effectiveness of prediction-based self-supervised learning in computer vision tasks, this paper explores the application of the MIM self-supervised learning framework in the field of driver distraction behavior detection Instead of using traditional supervised learning for model training and learning, we proposed a combined approach of pre-training and transfer learning. This approach effectively addressed the low generalization and substantial human consumption of labeled data in existing supervised learning methods. Instead of using convolutional neural networks, the Swin Transformer [28] is used as the encoder in the framework, taking advantage of the multi-head self-attention mechanism to overcome the insufficient global feature-capturing ability of existing CNN models. A large and high-dimensional classification dataset is used for pre-training, which provides a strong foundation for model representation learning and also makes the transferred model more robust and capable of better generalization. This will lead to better adaptability for recognition in various driving scenarios.

The main contributions of our work are summarized as follows:

(1) In order to address the high research cost and consumption issues associated with current supervised learning methods, we introduce a self-supervised learning method based on MIM for the detection of driver distraction behavior, named SL-DDBD. This will alleviate the time and effort consumption of data labeling. Meanwhile, through transfer learning, the model becomes more adaptable in downstream task detection scenarios, compared to the original supervised model, it has more effective feature attention and strong generalization ability.

(2) In this work, we utilize the currently more performant Swin Transformer as the encoder for the self-supervised learning framework. In the transfer learning process, we consider the depth of the Swin Transformer structure and its relationship with the application scenario. Taking into account the features of the driver distraction detection task, the number of Swin Transformer blocks in stage 3 of the encoder structure is reduced to effectively decrease the computational cost and parameters of the massive encoder structure. At the same time, in order to balance the computation of the multi-head self-attention [29] mechanism module and the improvement of the detection performance, the number of heads for W-MSA and SW-MSA [28] detection in each stage is re-distributed. This adjustment alleviates the computational redundancy of the encoder in more detail while providing better recognition for driver distraction behavior detection.

(3) In the model training, the best masking strategy was

applied to improve the detection performance, and the proposed data augmentation strategies in the paper were used for learning and training. Through experiments, a masking strategy with patch size of 64 and ratio of 0.5 was selected, which resulted in a more accurate recognition performance. A multi-strategy data augmentation approach was introduced in this work, including color jitter, motion blur, Gaussian noise, Mixup [30], and Cutmix [31] to better simulate a more diverse and realistic driving scenario, and improve the diversity of images. This enhances the robustness and generalization ability of the model.

(4) This paper compares the performance of self-supervised learning models and supervised learning models on the same dataset and training configuration through visual comparison. The results further validate that self-supervised models are better at capturing high-level semantic information and have more focused feature attention on critical discrimination regions. The work also compares its results with the state-of-the-art supervised and unsupervised learning methods in driver distraction behavior detection, demonstrating the advanced and feasibility of the proposed method.

## II. RELATED WORK

In recent years, there has been significant progress in related research fields such as CNN [32] [33] [34] [35], attention mechanism [36] [37] [38], knowledge distillation [39] [40] [41] and Transformer [42]. Depending on the advances in these related fields, researchers have combined these methods with research areas. This has led to a wide range of developments in driver distraction behavior detection.

### A. Driver Distraction Detection

Driver distraction detection is a hot topic of ADAS. Many traditional methods detect driver distraction by integrating CNN or graph neural network (GNN) models and have achieved good results. Hu et al. [32] proposed MVCNet, which processes information from three visual contents, including texture features, and incorporates GNN and CNN, optical flow, and semantic segmentation information. They also created a new semantic attention module that is integrated into both the CNN and graph attention network (GAT) branches. The multi-branch model decodes the integrated features to finally generate the driver's gaze attention map. Different from previous work, Xing et al. [33] designed a system that can simultaneously recognize a driver's body activities and mental state. The system is based on a framework of deep encoders and decoders. The encoder is designed based on CNN to extract effective spatial information from driving activity videos, while the encoder is designed using a fully connected network and LSTM-based RNN for estimating different driving states.

The areas of computer vision have extensively used attention mechanisms, which have improved the overall performance of networks and greatly helped with feature extraction and classification. Huang et al. designed a deep 3D network (D3DRN-AMED) for driver distraction detection. They inserted the attention mechanism as a non-linear transform layer into the residual network with a soft threshold to extract features related to driver behavior. Li et al. [37] used an object detection method to implement recognition of driver behavior and proposed an AB-DLM network model for driver distraction detection. They stacked SE attention mechanism module in the architecture and employed bi-directional feature pyramid networks (BiFPN) instead of the original path aggregation network (PANet) as a new multi-scale fusion network.

On the other hand, the lightweight of driver distraction detection model is very important for deployment to mobile devices. Some researchers have shifted their focus to knowledge distillation. Liu et al. [39] created a high-performance teacher network. Knowledge distillation was then used to direct the student network's learning process after gradually enhancing CNN's robustness to illumination shifts from shallow to deep layers. The distilled knowledge is transferred from the teacher network to the student network, resulting in a student network with 2.03M parameters. This is a powerful method for lightweight the driver behavior recognition model. Transformer is also gradually growing in the computer vision field. However, most researchers have not yet done much work in driver behavior detection. Wang et al. [42] proposed a detection model that integrates both CNN and Transformer architecture, incorporating specific enhancements to the Transformer structure for better performance. The original multilayer perceptron (MLP) module is replaced with a convolution module to reduce the number of model parameters and improve detection speed. A label-smoothed loss function is designed and applied to the model learning.

### B. Self-supervised Learning for Masked Image Modeling

Initial study and experimentation on self-supervised learning was conducted in the area of natural language processing before gradually expanding to computer vision. [43]. Nowadays, self-supervised learning using masked image modeling has slowly gained attention [44] [45] [46] [47] [48].

Bidirectional encoder representation from image transformers (BEiT) [44] is a pioneering work that transfers the BERT-style pre-training method to the visual domain and introduces the concept of MIM pre-training, making important contributions to the field of self-supervised learning. BEiT first annotates the original image and randomly masks image blocks, feeding the masked image into the encoder, with the main pre-training goal being to recover the masked image blocks based on the unmasked ones. Masked autoencoders (MAE) proposed by He et al. [45] is a highly influential work that drives the development of MIM self-supervised learning. MAE compared to BEiT, advocates a simpler training logic. It first proposed random masking of images and directly reconstructed masked image blocks for training. Adopted an asymmetric structure of encoder-decoder, where the encoder only calculated non-masked image blocks and used a lightweight decoder design. Soon after MAE was proposed, Xie et al. [46] proposed a simpler mask learning framework. Regressing to a more original and simple design, including random masking, regressing RGB pixel values, and directly using a single linear prediction head. A simple framework for masked image modeling (SimMIM) also achieved good results, achieving a top accuracy of 87.1 on ImageNet-1k. Later, Chen et al. [47] noticed some shortcomings in previous work, such as a focus on semantic learning in image center blocks and neglect of edge regions, which also made context autoencoder (CAE)'s contribution one of the models not only focus on the center area of the image. Secondly, CAE splits representation learning and front-end tasks, making the encoder more focused on representation learning.

The UDL model proposed by Li et al. [49] used a traditional contrastive unsupervised learning method. In the UDL model, a new backbone and projection head were constructed using MLP. They also designed and used a loss function with a stop-gradient strategy to guide learning and training, resulting in a more robust model. Although this was a good attempt, research on self-supervised learning with masked image modeling method in the field of driver behavior detection is still scarce.

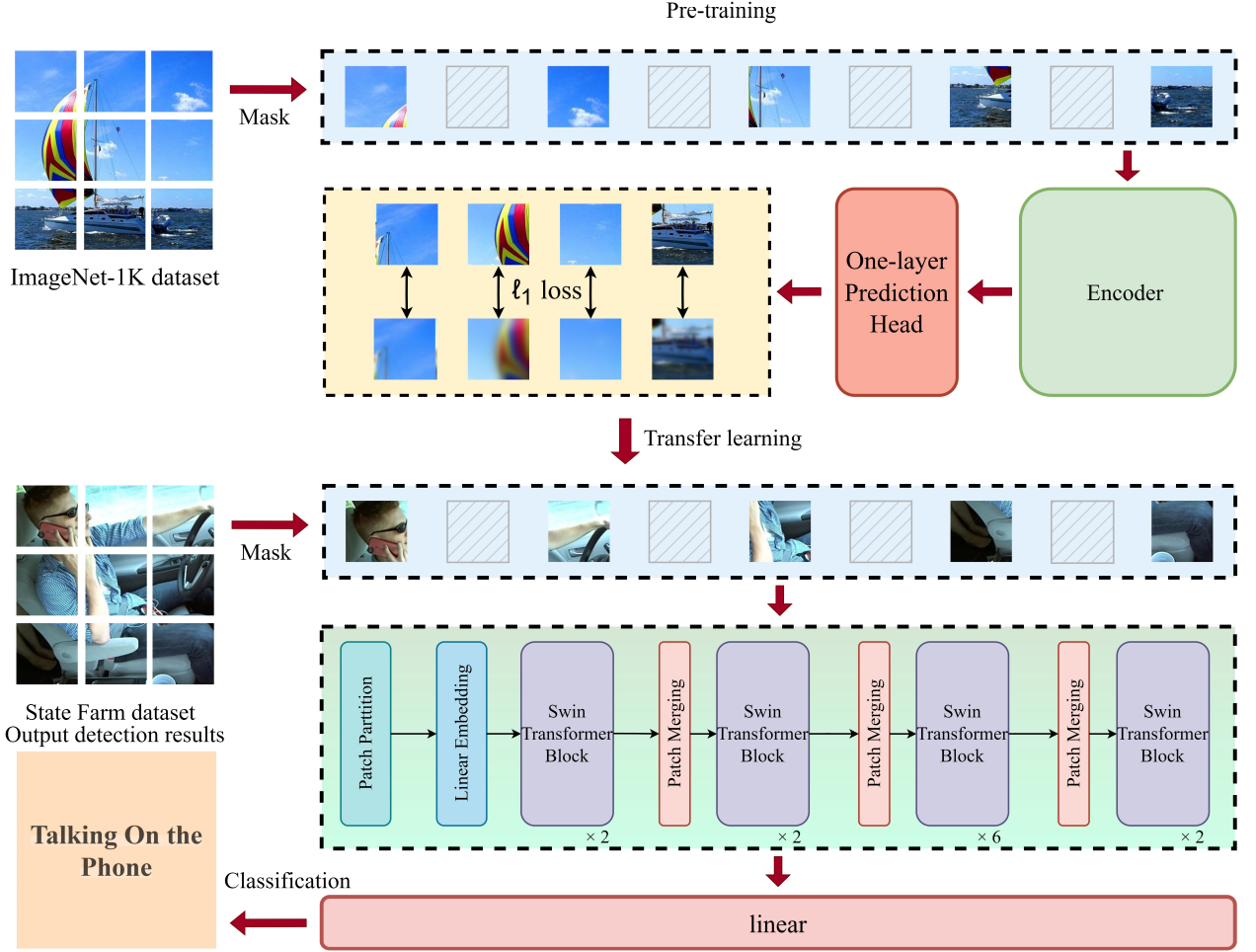

**Fig. 1.** The pipeline of the proposed SL-DDBD.

## III. METHODS

### A. A Suitable Self-supervised Learning Framework

Subsequently, we propose a self-supervised learning method based on masked image modeling for detecting driver distraction behavior. The proposed method consists of two stages: pre-training and fine-tuning. In the pre-training stage, a large unlabeled dataset is used for self-supervised learning. The masked image modeling method is employed to mask a part of the input image and train the model to re-predict the masked area of the image. The fine-tuning stage uses transfer learning to fine-tune the model with a specific dataset. This stage includes precision optimization and lightweight encoder, as well as data augmentation. The best masking strategy is determined through experiments. The fine-tuned model is capable of adapting to downstream tasks with better adaptability. The self-supervised learning pre-training framework consists of four main parts:

(1) Masking strategy: This section focuses on how to choose the masked part of the image and considers the overall level of masking and the size of the masked image block. The output is then passed to the next section.

(2) Encoder architecture: A strong encoder with good learning ability can be used for various visual downstream tasks. The encoder extracts the latent representation of the masked image and predicts the original signal of the masked part through training and learning.

(3) Prediction head: The prediction head is used for the latent representation to represent the original signal of the masked patch.

(4) Prediction target: This part specifies the form of the original signal to be predicted, which can be either the original pixel values or the transformation of the original pixels. The loss type is also defined in this part. Cross-entropy categorization loss and L1 or L2 regression loss are common choices.

In the masking strategy, a random masking strategy is used. Image blocks are the basic processing units in the entire system framework. Using this as a basic unit can facilitate masking operations at the block level. It is convenient and easy to implement masking, complete masking, or no masking on the customizability of the image. For different encoders, there are different patch sizes. For the Swin Transformer encoder, we consider patch size dimensions at different resolution levels, ranging from $4 \times 4$ to $32 \times 32$. For the ViT encoder, a default patch size of $32 \times 32$ is used.

In the prediction head, it is necessary to ensure that the input to the prediction head is consistent with the output of the encoder. Once the prerequisites are met, the form and size of the prediction head can be customized. The prediction head is then defined to predict the target. In some early works, an autoencoder was followed with a heavy prediction head. The use of a complex detection head did not result in better performance but rather increased the training cost. In this paper, considering the application scenario of driver distraction detection, a simple prediction head is used to accomplish this task. We use $1 \times 1$ convolution kernels to implement a single linear layer to predict pixel values.

In the prediction target, the pixel values are continuous in the color space, and the original pixels of the masked area are predicted directly through regression. Each feature vector in the feature map is mapped back to the original resolution and is in charge of forecasting the original pixels. This is used to predict all the pixel values of the input image at all resolutions.

We apply a $1 \times 1$ linear layer with an output dimension of $3072 = 32 \times 32 \times 3$ to the $32 \times 32$ down-sampling feature map made by the Swin Transformer encoder to depict the RGB values of the $32 \times 32$ pixels. In order to account for lower-resolution objects, the original image is down-sampled at multiple dimensions, respectively. These include $\{32 \times, 16 \times, 8 \times, 4 \times, 2 \times\}$.

Use $\ell 1$ - loss on the masking pixels:

$$L = \frac{1}{\Omega(x_M)} \left\| y_M - x_M \right\|_1 \tag{1}$$

where $x$, $y \in \mathbb{R}^{3HW \times 1}$ are the input RGB and predicted values, $M$ is the collection of masked pixels, $\Omega(\cdot)$ is the number of elements.

### B. Encoder Architecture

Due to the high requirements of the encoder to extract the latent feature representation of the masked part of the image, the encoder needs strong representation learning ability. Therefore, we consider using the Swin Transformer as the encoder in the self-supervised learning framework.

Transformer has shone in the field of computer vision after being transferred from the NLP field, gradually becoming the main general pillar of computer vision. However, After ViT, the birth of Swin Transformer was a milestone that pushed the development of Transformer for computer vision tasks [50]. Swin Transformer is more suitable for visual tasks, with two important design points: hierarchical Transformer and shifted window, resolve large scale differences in visual entities and the high pixel resolution of images compared to text.

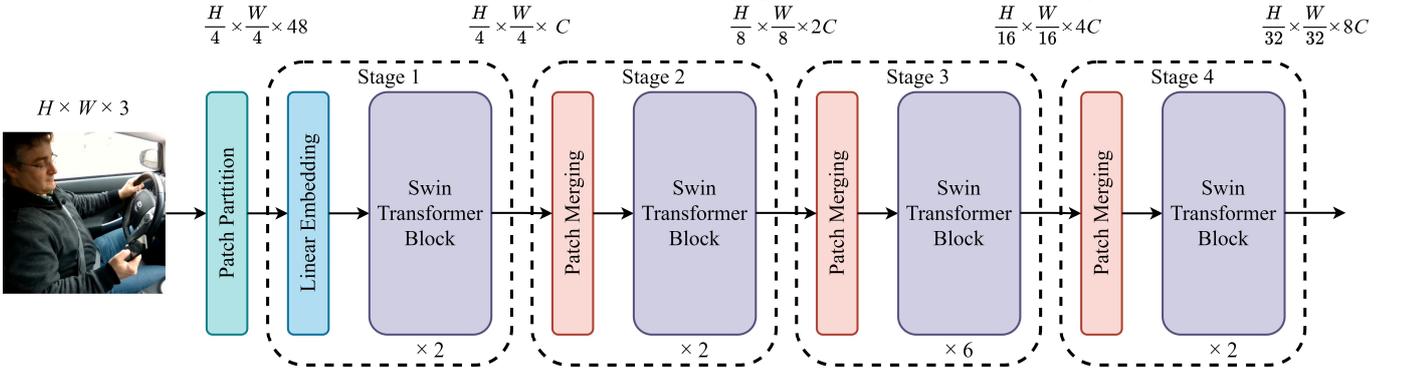

**Fig. 2.** Architecture of encoder in the self-supervised learning framework.

The architecture of encoder as shown in Fig. 2. First, the image is input into the Patch Partition module for block processing. The image is divided into multiple patches of $4 \times 4$ pixel size. Then, it is flattened in the channel direction. When an RGB three-channel picture is used as the input, then each patch has $4 \times 4 = 16$ pixels, and each pixel has R, G, and B three values, so after flattening, it is $16 \times 3 = 48$. After processing by the Patch Partition, the image shape changes from $[H, W, 3]$ to $[\frac{H}{4}, \frac{W}{4}, 48]$. The Linear embedding layer linearly transforms each pixel channel, that is, from the original

$48$ to $C$ . $C$ value is defined. After this layer, the image shape is $[\frac{H}{4}, \frac{W}{4}, C]$.

The image is then processed through four stages one by one and the size changes accordingly. Linear embedding layer is only in stage 1. Patch merging layers and various numbers of Swin Transformer blocks make up the final stages. The Swin Transformer block, as the main working structure, contains two structures. One is the block using the W-MSA module, and the other is the block using the SW-MSA module. So when stacking the Swin Transformer blocks, they are stacked in pairs.

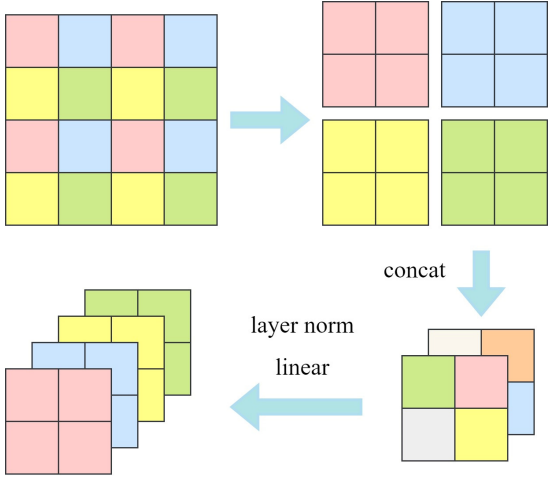

**Fig. 3.** The workflow of Patch Merging.

The patch merging process is shown in Fig. 3. When the input is a single-channel feature map of $4\times4$ size, Patch merging divides each $2\times2$ neighboring pixel into a patch. By connecting the pixels in the same location in each patch, four feature images are created. Concatenating four feature maps in the depth direction, followed by passing through a LayerNorm layer. Finally, a fully connected layer performs linear transformation on the depth direction of the feature map, reducing the depth of the feature map from $C$ to $\dfrac{C}{2}$. As a result, after stages 2, 3, and 4, the shape of the image will be halved in both width and height while the number of channels will be doubled.

### C. Improved Data Augmentation Strategy

In terms of the dataset, it is dangerous to create driving scenarios that simulate actual driver distraction. Therefore, the number of real driver distraction datasets is relatively small, and the quality of the datasets still needs to be improved. The quantity and diversity of the dataset for deep learning network models often greatly affect the training results of the model, including the recognition accuracy and robustness. To tackle this problem, this work presents a data enhancement strategy that aims to simulate and closely mimic real-world driving scenarios to increase the quantity and diversity of the dataset.

(1) Color jitter: An effective data augmentation method, randomly changing the exposure, saturation, and hue of the images. The aim is to simulate the driving situations of the driver in the driving room under different lights and weather conditions. This increases the number and diversity of training datasets, allowing the model to learn the differences brought by changes in lighting.

(2) Motion blur: Turns a clearly focused image into a motion blur effect. Most of the images are captured in focus for easier feature extraction. In realistic conditions, the driver's movements are fast and continuous in the video of the detection device. Therefore, it is necessary to consider the phenomenon of motion blur caused by the driver continuously performing actions.

(3) Gaussian noise: A form of noise whose probability density function has a Gaussian or normal distribution is referred to as

Gaussian noise. It applies overall noise to images in order to enhance their diversity and improve the model's ability to learn representations.

(4) Horizontal flipping and random scaling: Random scaling enhances the model's ability to detect distracted driving actions of multiple scales, while horizontal flipping better simulates the driving scenarios of right-side and left-side vehicles.

(5) Cutmix: A random crop box is generated, and a corresponding portion of the A image is cropped. Then the corresponding ROI in the B image is placed in the cropping area of the A image to form a new image. The loss is also solved by weighted summing [51]. Employing hard fusion of two images and concurrently implementing soft fusion strategies for labels, Cutmix ensures the distribution of the dataset remains unaltered. Let $x \in \mathbb{R}^{W\times H\times C}$ and $y$ represent the training images and their labels. The merge operation is defined as:

$$\tilde{x} = \mathbf{M} \odot x_A + (\mathbf{1}-\mathbf{M}) \odot x_B \qquad (2)$$

$$\tilde{y} = \lambda y_A + (1-\lambda) y_B \qquad (3)$$

where $\mathbf{M} \in \{0,1\}^{W\times H}$ represents a binary mask indicating the deleted and filled positions in the image. $\mathbf{1}$ is a binary mask filled with 1, and $\odot$ represents element-wise multiplication. $\lambda$ belongs to the $Beta(\alpha,\alpha)$ distribution, and if $\alpha=1$ is set in the experiment, $\lambda$ follows a uniform distribution between $(0,1)$.

To sample the binary mask $\mathbf{M}$, it is necessary to sample the bounding box $\mathbf{B}=(r_x,r_y,r_w,r_h)$ of the cropping area. Then, sample $x_A$ and $x_B$ based on the sampling result for cropping and padding. In the experiment, the aspect ratio of the rectangular mask $\mathbf{M}$ is proportional to that of the original image. The bounding box coordinates are sampled as follows:

$$r_x \sim \mathrm{Unif}\,(0,W), r_w = W\sqrt{1-\lambda} \qquad (4)$$

$$r_y \sim \mathrm{Unif}\,(0,H), r_h = H\sqrt{1-\lambda} \qquad (5)$$

The cropping area ratio is $\dfrac{r_w r_h}{WH}=1-\lambda$, $\mathbf{M}=0$ in the cropping area B, while the rest of the area has $\mathbf{M}=1$.

(6) Mixup: A regularization technique that randomly blends the pixels of two training images to create a new image that incorporates the labels of both input images [52]. This approach is designed to enhance the diversity and complexity of the training dataset by maximizing the combination of different contextual information, thereby improving performance. The calculation formula is as follows:

$$\tilde{x} = \lambda x_i + (1-\lambda)x_j \qquad (6)$$

$$\tilde{y} = \lambda y_i + (1-\lambda)y_j \qquad (7)$$

where $\lambda \in [0,1]$ follow a $Beta(\alpha,\alpha)$ distribution, $\alpha$ is a hyperparameter that controls the interpolation strength. The larger the value of $\alpha$, the more obvious the interpolation effect. While the smaller the value, the closer the Mixup enhancement effect tends to be ineffective.

## D. Self-Supervised Pretraining

In self-supervised learning, the training process is similar to a conventional autoencoder. In this work, the input information is mapped to latent features representation using a Swin Transformer encoder. Then the liner layer is used as a prediction head to reconstruct the masked part of the image by potential feature information. During pre-training, large-scale, unlabeled datasets are used for representation training and learning. In this work used ImagenNet-1K dataset, which contains no labels. The images in the dataset are subjected to a random masking strategy in the self-supervised learning framework, with patch size set to 32 and ratio set to 0.5. The inputs are fed into the encoder, which processes both the visible tokens and the masked tokens.

The encoder's core working module is the Swin Transformer block. In the Transformer, the self-attention module is the basic operational unit, just like the convolutional operation in a CNN network. The self-attention mechanism in the Transformer block is able to adaptively model the long-term dependency relationships among sequence markers. One type of dependency relationship can only be established by a particular attention function. Multi-head self-attention (MSA) aims to learn multiple dependencies from different representation subspaces. In particular, the keys, values and queries of the $d_{model}$ dimension are divided linearly into $h$ groups, the attention function is carried out in all groups concurrently, generate $d_{model} / h$ dimension output values and these values are concatenated and projected.

MSA is expressed as follows:

$$\text{MultiHead}(\boldsymbol{Q}, \boldsymbol{K}, \boldsymbol{V}) = \text{Concat}(y_1, \ldots, \boldsymbol{y}_n) W^O \quad (8)$$

$$y_i = \text{softmax}\left(\frac{Q_i K_i^T}{\sqrt{d_k}}\right) V_i \quad (9)$$

where $d_k = d_{model} / h$ is the average feature area of each attention head. $Q_i = X_i W_i^q$, $K_i = X_i W_i^k$, and $V_i = W_i^v$ stand for the key, query and value. $W^O$ is the learnable projection matrix, and $h$ is the total number of self-attention head. $W_i^q, W_i^k, \ W_i^v$ are parameter matrices and $X_i$ represents the feature matrix of the $i$-th head.

## E. Transfer Learning With Accuracy Optimization and Light Weighting

In the field of computer vision, large networks are designed to provide better service for high-difficulty image classification tasks and object detection. However, these deep and wide networks also drive the progress of various visual downstream tasks. Transfer learning is an important part of realizing downstream tasks [45]. We consider the relationship between complex networks and the specific application scenario in this work. In order to achieve better detection results, we also need to consider the deployment of the model to hardware and specific engineering applications. Therefore, we perform lightweight and precision optimization tuning work on the encoder in the self-supervised learning framework.

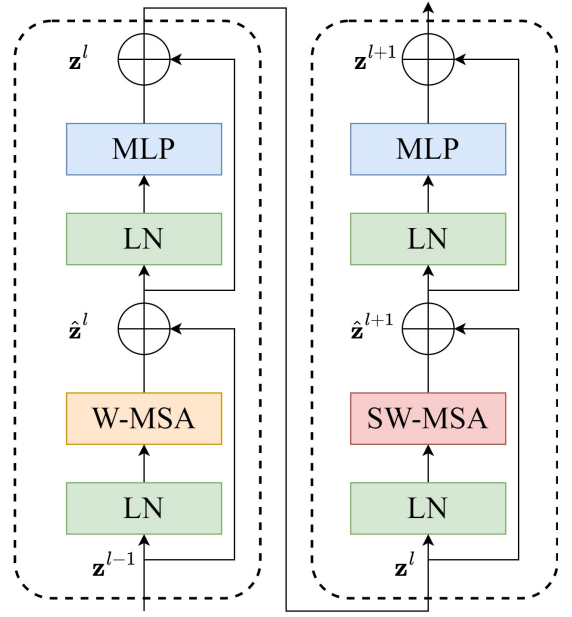

**Fig. 4.** The network of Swin Transformer block

The Swin Transformer block is shown in Fig. 4. The image input enters the first block, where the image needs to go through a layer norm layer and a W-MSA module, with a skip connection alongside both processing steps. The image continues to enter the layer norm layer and MLP module, with a skip connection alongside this path as well. At this point, the image has completed processing through the first block and outputs to the second block. The second block is similar in overall structure to the first, but instead of using the conventional W-MSA, it uses SW-MSA. The consecutive Swin Transformer block calculation process is as follows:

$$\hat{\boldsymbol{z}}^l = \text{W-MSA}\left(\text{LN}\left(\boldsymbol{z}^{l-1}\right)\right) + \boldsymbol{z}^{l-1} \quad (10)$$

$$\hat{\boldsymbol{z}}^l = \text{MLP}\left(\text{LN}\left(\hat{\boldsymbol{z}}^l\right)\right) + \hat{\boldsymbol{z}}^l \quad (11)$$

$$\hat{\boldsymbol{z}}^{l+1} = \text{SW-MSA}\left(\text{LN}\left(\boldsymbol{z}^l\right)\right) + \boldsymbol{z}^l \quad (12)$$

$$\hat{\boldsymbol{z}}^{l+1} = \text{MLP}\left(\text{LN}\left(\boldsymbol{z}^{l+1}\right)\right) + \hat{\boldsymbol{z}}^{l+1} \quad (13)$$

where $\hat{\boldsymbol{z}}^l$ and $\hat{\boldsymbol{z}}^{l+1}$ denote the output features of the (S)W-MSA module and MLP module of block $l$, respectively; W-MSA and SW-MSA represent the multi-headed attention from the partitioned configuration with regular window and shifted window.

In the encoder, each image goes through W-MSA and SW-MSA processing in Swin Transformer block. Therefore, the number of Swin Transformer blocks must be even. Compared to large and difficult visual tasks such as world object classification, driver distraction behavior classification is less complex and has more distinctive features. Although large networks perform well in recognition of complex classification tasks, overly complex networks are not the best solution for relatively simple detection scenarios. In order to reduce the computation and parameters of the massive encoder, the work changes the number of stacked Swin Transformer blocks in stages of the encoder structure. Specifically, the original 18 stacked Swin Transformer blocks in the third stage are reduced to 6.

The W-MSA and SW-MSA are the main working modules. Compared to the original MSA, the Swin Transformer block performs multi-head self-attention on the window. The MSA, due to its window-based setting, significantly reduces the computation cost compared to the original. The computation cost formula of MSA is as follows:

$$\Omega(\text{MSA}) = 4hwC^2 + 2(hw)^2 C \tag{14}$$

where $\Omega$ is the computation, $h$ and $w$ are the height and width of the image respectively, and $C$ is the number of channels. W-MSA module divides the feature map into a window with the width and height of M. A feature map that will get $\dfrac{h}{M} \times \dfrac{w}{M}$ windows, and then use the multi-headed self-attention module for each window. Since the window's width and height are $M$, bring the above formula as $4(MC)^2 + 2(M)^4 C$, the final W-MSA calculation is:

$$\Omega(\text{W-MSA}) = 4hwC^2 + 2M^2 hwC \tag{15}$$

It is similar that the number of heads in the MSA is controllable and the size of the encoder is positively correlated with the number of heads. The number of heads in the W-MSA and SW-MSA in the four stages of the encoder are set differently. The original baseline sets the number of heads to {4, 8, 16, 32}. However, in the case of very few categories, the redundant number of heads brings more computational parameters and does little to enhance the detection task for fewer categories. Hence, we adjust the number of detection heads in each stage to {3,6,12,24}.

## IV. EXPERIMENTS

### A. Datasets and Comparison of Each Data Augmentation Strategy

The dataset used in this work is the State Farm dataset from the official Kaggle competition [53]. This dataset is a comprehensive and diverse dataset for driver behavior monitoring, which includes 26 participants of different races, skin colors, and genders (13 male and 13 female) from America, Asia, and Africa. All images in the dataset were captured by a camera fixed in the car dashboard, and all images are RGB pixels. The dataset consists of a total of 22424 images. As our initial dataset, we randomly divided each class of images into 80% for training and 20% for testing.

We used the proposed data augmentation strategy to enhance the dataset to improve the generalization ability and robustness of the proposed detection algorithm model. After data augmentation, the training images are 81976 and the test images are 20,000. The specific classification of distracted driver behavior in the State-Farm dataset is shown in Table Ⅱ. The basic dataset division is shown in Table I (a), and the division of the expanded dataset using our proposed data augmentation strategy is shown in Table I (b).

TABLE I
DATASET DIVISION

|     |       | C0 | C1 | C2 | C3 | C4 | C5 | C6 | C7 | C8 | C9 |
|-----|-------|------|------|------|------|------|------|------|------|------|------|
| (a) | Train | 1991 | 1814 | 1854 | 1877 | 1861 | 1850 | 1860 | 1602 | 1529 | 1703 |
|     | Test  | 498 | 453 | 463 | 469 | 465 | 462 | 465 | 400 | 382 | 426 |
|     | Total | 2489 | 2267 | 2317 | 2346 | 2326 | 2312 | 2325 | 2002 | 1911 | 2129 |
| (b) | Train | 9156 | 8268 | 8468 | 8584 | 8504 | 8448 | 8500 | 7208 | 6844 | 7996 |
|     | Test  | 2000 | 2000 | 2000 | 2000 | 2000 | 2000 | 2000 | 2000 | 2000 | 2000 |
|     | Total | 11156 | 10268 | 10468 | 10584 | 10504 | 10448 | 10500 | 9208 | 8844 | 9996 |

In order to explore the positive effects of the data augmentation strategy proposed in this work on driver distraction detection, we evaluated the effectiveness of each data augmentation strategy. This exploration aims to discover the potential connection between different data augmentations and driver distraction detection. Under the same software and hardware environment, we set the unified masking strategy of patch size 64, ratio 0.5, and used the improved encoder. We trained seven models with each of the six data augmentation strategies and without any data augmentation. The performance of the seven models was tested on the same dataset and the results are shown in Fig. 5.

The NoAug model (without any data augmentation strategy) has the worst overall performance, with the lowest precision in multiple categories, including C0, C1, C3, C6, C7, and C8. The models with Cutmix and Mixup perform the best among all data augmentation strategies, followed by HFRS (Horizontal flipping and random scaling) and Motion blur, and finally Gaussian noise and Color jitter. The precision of the models improved significantly after data augmentation. For example, the NoAug model has the lowest recognition precision in category C8, but the precision in this category for all other models augmented has improved. The precision of the Motion blur and Gaussian noise models is slightly degraded on C2 and C4, both of which belong to the "talking on the phone" category. Motion blur and Gaussian noise affect the images that have few phone pixels in the image, so there is a small decrease in precision in both categories. However, the overall accuracy is still higher than that of NoAug model. Since the Cutmix and Mixup data augmentation methods remove and overlap some parts of the image, causing significant changes in the image. These data augmentation strategies increase the training difficulty but also improve the model's feature extraction from the image, resulting in highly robust and accurate models after training. Next in line for effectiveness are the HFRS and Motion blur. Like the previous data augmentation strategies, HFRS also makes significant changes to the image by randomly altering its size and scale, thus enhancing effective feature extraction. The improvements in precision brought by Gaussian noise and Color jitter are not obvious because they only induce simple changes to the image by adding noise and color variations. The proposed data augmentation strategy in this work has a positive effect on the detection of distracted driver behavior. It enhances the diversity of the dataset, achieves data expansion, and improves model detection precision.

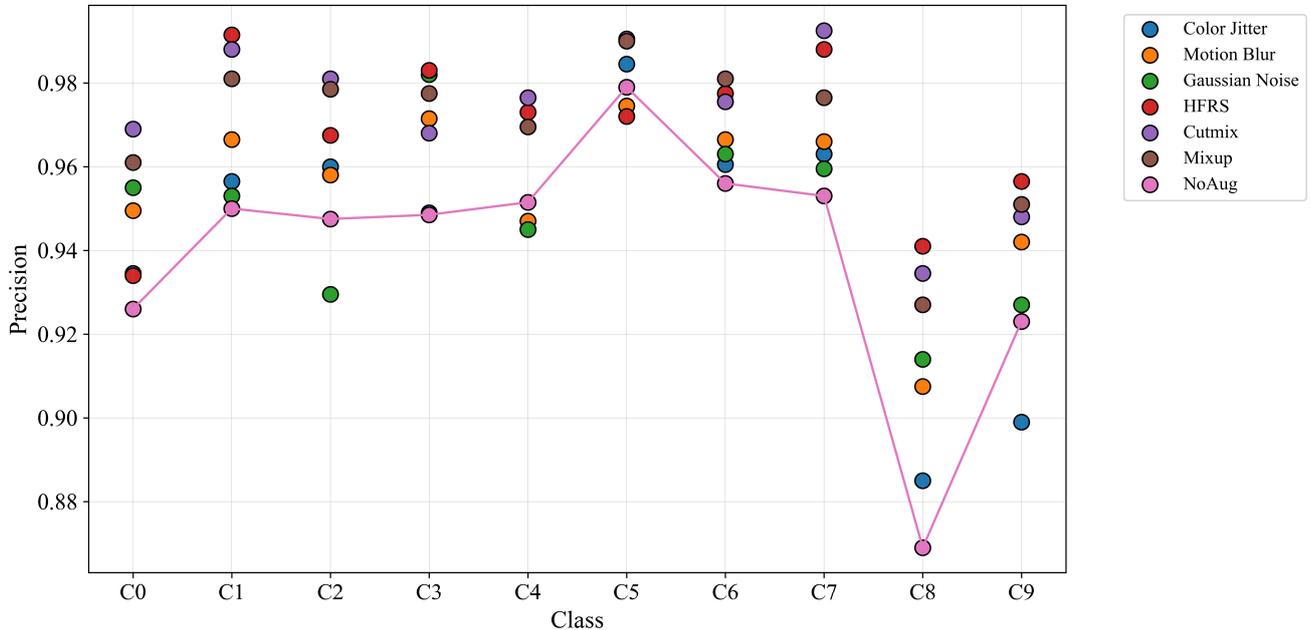

**Fig. 5.** Comparison of precision of models using different data enhancement strategies.

TABLE II
THE DRIVER BEHAVIORS IN THE STATE-FARM DATASET

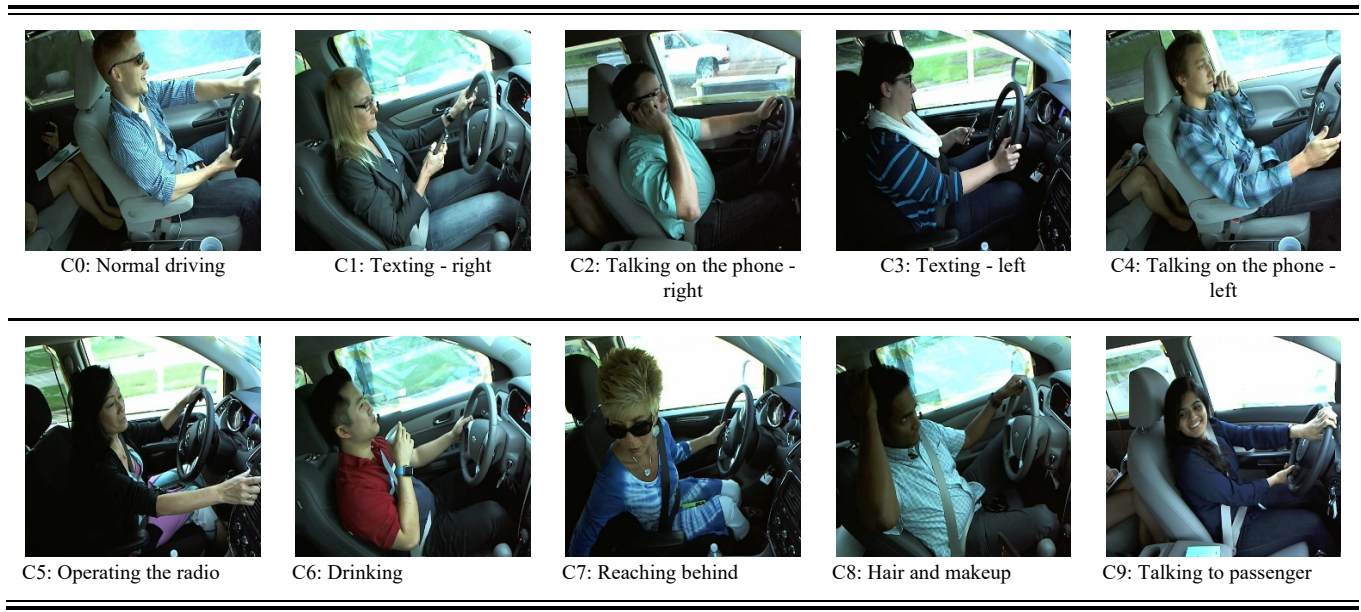

### B. Training Details

This paper uses the experimental environment configuration in Table III to guarantee the effectiveness of model training and testing. A large-scale self-supervised pre-training is conducted using the ImageNet-1k dataset. During pre-training, the size of input images is adjusted to $192 \times 192$, and the window size is adjusted to 6 to adapt to the changed input image size. In self-supervised pre-training, the training cycle is 800 epochs, and AdamW optimizer with cosine learning rate adjustment is used. The specific training hyperparameters are as follows: batch size is 2048, $\beta1 = 0.9$, $\beta2 = 0.999$, weight decay is 0.05, and the base learning rate is 8e-4. In the masking strategy for masked image modeling: the random masking strategy is used, the masked ratio is 0.5, and the patch size is 32. The predicted target image size in the linear prediction head is $192 \times 192$.

TABLE III
HARDWARE AND SOFTWARE CONFIGURATION.

| Operating systems | Linux Ubuntu 20.04.2 LTS |
|---|---|
| CPU | Intel Xeon E5-2680 v4 |
| GPU | NVIDIA Geforce RTX 2080 ti（12G） |
| Memory | 16G × 2 |
| Solid state drives | 480G |
| Pytorch | V1.10 |
| CUDA | V10.0 |

In the transfer learning process, we used both the basic version and the augmented version of the Sate-Farm dataset. In the hardware and software environment, we used six Intel Xeon E5-2680 v4 CPUs and an NVIDIA RTX 2080 TI GPU with a 10G memory size. We also used the AdamW optimizer for transfer learning, with all training periods changed to 110 epochs, the batch size adjusted to 32, the base learning rate changed to 5e-3, and the weight decay remaining at 0.05. In the masking strategy for image modeling, we used a random masking strategy. The masked ratio and patch size were used as variable parameters in the experiment. The image input size was uniformly changed to $224 \times 224$, and the window size was adjusted to 7.

### C. Self-supervised Learning and Masked Image Modeling Visualization

To investigate the advancement of self-supervised learning based on masked image modeling, we visualize the self-supervised learning model and the supervised learning model, and the results are shown in Fig. 6. We adopt the same encoder structure, training settings, and dataset. We trained a supervised learning model. The left column shows the original images of the driver distraction behavior dataset, the middle column shows the visualization effects obtained by the supervised model, and the right column shows the visualization effects obtained by the self-supervised model in this work. We use the grad-cam method [54] for visualization and use gradients for visualization. We select the Layer-Norm layer of the last Swin Transformer block module in backbone.

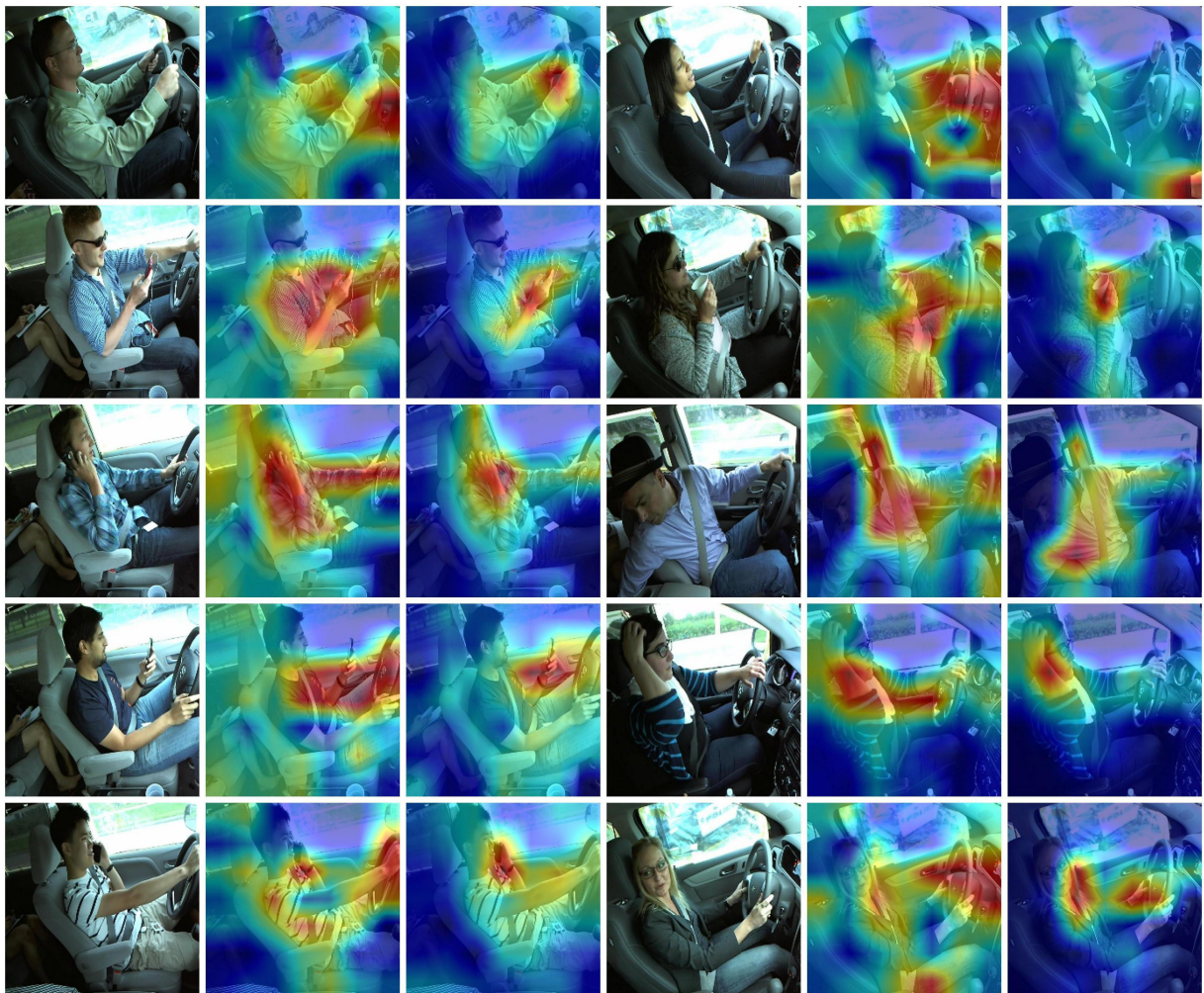

**Fig. 6.** Self-supervised and fully supervised learning visualizations on State Farm dataset.

The visualization comparison results in the 10 types of driver distraction show that the supervised learning model's attention is relatively dispersed in all categories. For example, using a cell phone for calling and texting, the supervised model overall focuses on a large area. At the same time, it also sometimes needs to pay attention to the situation on the steering wheel to distinguish distracted behavior. However, the self-supervised learning model has a very small focus area and concentrates on the key determining parts of the driver's behavior. This is particularly evident in the category of the driver drinking, where the self-supervised learning model focuses only on the driver's hand holding the cup in front. The same is true for distracted behavior using a phone; where only the hand and the phone part need to be focused on to classify the behavior. Overall, the self-supervised model's attention is focused on the key parts of the scene objects and has a better grasp of the feature information, avoiding the problem of feature redundancy and excessive computational costs.

*D. Comparison of Transfer Learning Optimization*

This experiment investigates the impact of self-supervised masking strategies on detection performance by setting different masking details for the baseline. The patch size is the size of the masked image blocks in the masking strategy. After being masked, all patches are input into the encoder, so the patch size directly affects the encoder's receptive field. A larger patch size provides a larger receptive field but also contains more irrelevant features. Smaller patches reduce the receptive field and limit the communication between the overall image information. These situations affect the feature learning of the encoder and result in poor classification performance. The masked ratio is also an important part of the masking strategy. It affects the degree of masking of the entire input image. A smaller ratio may cause the encoder to learn weak representations and make it difficult to restore and predict the masked image. A larger ratio value leads to too much masking, making it difficult to train and resulting in a strong representation ability. Therefore, the ratio value will greatly impact the model's performance. The masking strategy is determined by the patch size and ratio value. In this experiment, we will set different ratio values and patch size values to verify the impact of the masking strategy on the model's detection performance.

In the experiment, set the patch size first, and then adjust the size of masked ratio value. The specific test results of different masking strategy models are shown in Table IV.

Acc represents the accuracy, and training time is the training time under the hardware conditions used. The baseline model's mask setting is patch size 32, ratio 0.5. It adopted the pre-trained model setting and encoder structure. The baseline accuracy is 84.92%. When the patch size is 32, the baseline is 0.31% higher than the masked ratio set to 0.4, and the masked ratio set to 0.6 is the highest accuracy of 88.86% at this size. When the patch size is 64, the masked ratio set to 0.6 has the highest accuracy of 88.70%. When the patch size and ratio value are larger, an accuracy of 88.70% can be achieved. This means that in a situation where the image is heavily masked and the model training is relatively difficult, the model has good learning representation ability. Even when the patch size is 16 and the masked ratio is set to 0.4, good performance results have been obtained. This means that in the case of less image masking, the model's feature learning for driver distraction detection scenes is favorable, and favorable recognition results have been obtained. The accuracy reached 89.33% when the masked ratio was 0.5. This is the highest accuracy obtained from the masking strategy experiment before optimization, compared to the baseline, it increased by 4.41%. The training time of all models fluctuated between 8:34:21 and 8:57:49. The longest training time did not exceed 9 hours. Therefore, overall, the change in patch size and masked ratio has a very small impact on the training cost of the entire model.

TABLE IV
Model Performance with Different Masking Strategy

| Patch size | Masked ratio | Acc(%) | Training time |
|---|---|---|---|
| 16 | 0.4 | 88.08 | 8:34:21 |
| 16 | 0.5 | 89.33 | 8:40:49 |
| 16 | 0.6 | 87.05 | 8:45:26 |
| 32 | 0.4 | 84.61 | 8:51:42 |
| 32 | 0.5 | 84.92 | 8:55:30 |
| 32 | 0.6 | 88.86 | 8:57:49 |
| 64 | 0.4 | 83.73 | 8:52:57 |
| 64 | 0.5 | 86.68 | 8:54:19 |
| 64 | 0.6 | 88.70 | 8:53:06 |

The experiment in this section will verify the impact of optimization in transfer learning on the model's detection performance. The optimized model will also experiment with different masking strategies. Fig. 7, Fig. 8, and Fig. 9 are the comparison results of the optimized model and the baseline in the case of patch size 16, 32, and 64 respectively. Each figure shows the accuracy of the two models under the same masking strategy, with the same patch size, presented in the form of a bar graph, making it more intuitive to show the accuracy improvement brought by the optimization work.

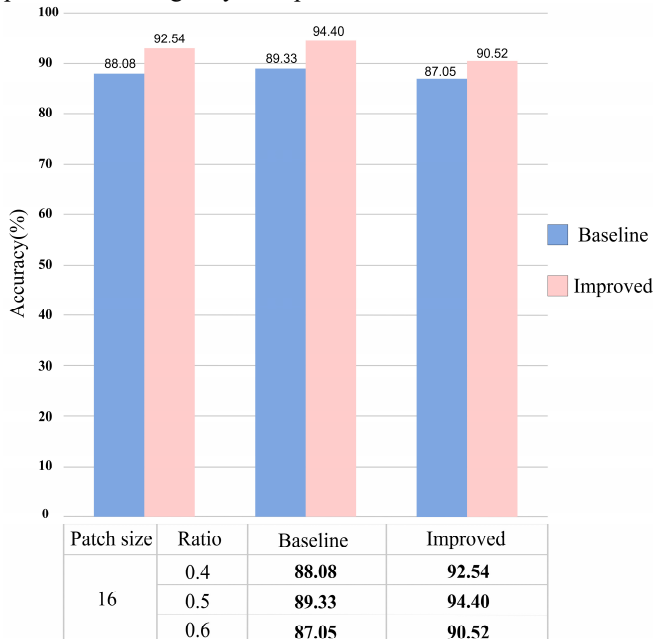

**Fig. 7.** The comparison of the accuracy of the baseline and improved models with a fixed patch size of 16 and different ratio values.

| Patch size | Ratio | Baseline | Improved |
|---|---|---|---|
| 16 | 0.4 | **88.08** | **92.54** |
| | 0.5 | **89.33** | **94.40** |
| | 0.6 | **87.05** | **90.52** |

In Fig.7, the patch size is 16 and all improved models increase accuracy by 4.46%, 5.07%, and 3.47% compared to the original model for each masked ratio. The original model has the lowest accuracy of 87.05% when masked ratio is 0.6. However, the optimized model still maintains an accuracy above 90% under this setting. At the same patch size setting, the optimized model reached an accuracy of 94.40% when masked ratio is 0.5, which is the highest accuracy under this setting.

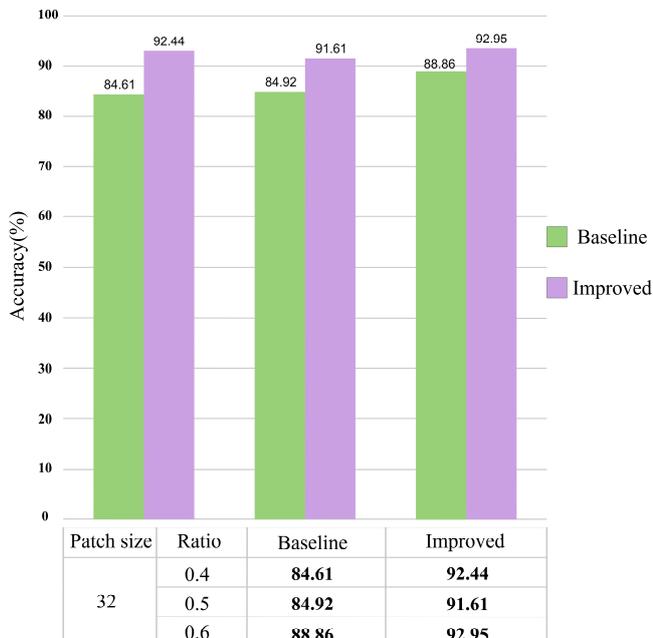

**Fig. 8.** The comparison of the accuracy of the baseline and improved models with a fixed patch size of 32 and different ratio values.

| Patch size | Ratio | Baseline | Improved |
|---|---|---|---|
| 32 | 0.4 | **84.61** | **92.44** |
| | 0.5 | **84.92** | **91.61** |
| | 0.6 | **88.86** | **92.95** |

The improved model in Fig. 8 shows an increase in accuracy of 7.83%, 6.69%, and 4.09% respectively compared to the original model. The baseline model has an accuracy of only over 84% when the masked ratio is 0.4 and 0.5. However, after the model is optimized, the accuracy improvement is significant, all reaching above 91%. Fig. 9 also shows that the performance of the improved model is still better than the original model. Especially when the masked ratio is 0.5, the accuracy of the improved model reaches 95.13%, which is the highest recognition accuracy among all masking strategies. Compared to the original model under the same strategy, it has increased by 8.45%, and compared to the baseline, it has increased by 10.21%. In all the above model comparisons, the performance of all improved models is better than the original model. The separate validation on various masking strategies indicates that the optimization work on accuracy in transfer learning is effective.

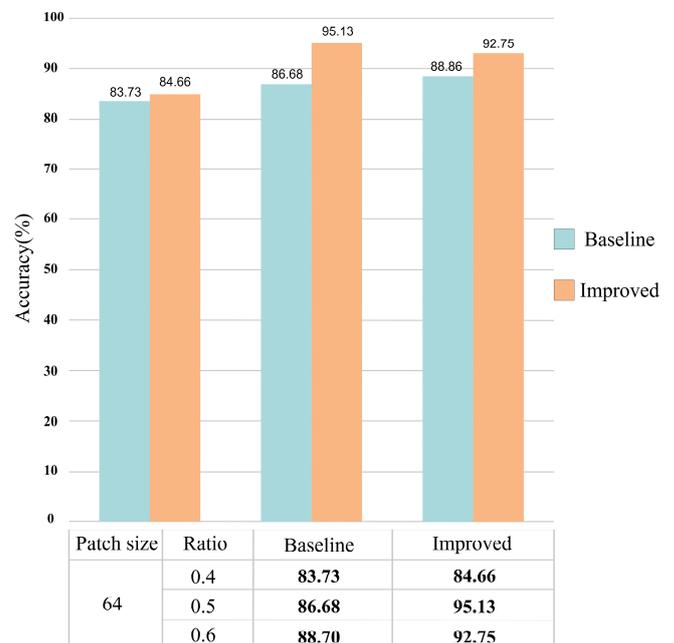

**Fig. 9.** The comparison of the accuracy of the baseline and improved models with a fixed patch size of 16 and different ratio values.

| Patch size | Ratio | Baseline | Improved |
|---|---|---|---|
| 64 | 0.4 | **83.73** | **84.66** |
| | 0.5 | **86.68** | **95.13** |
| | 0.6 | **88.70** | **92.75** |

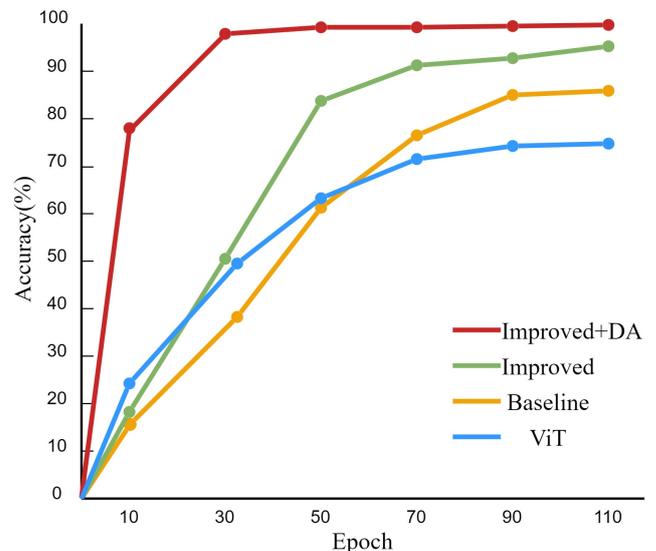

**Fig. 10.** Accuracy rate comparison of different models.

The optimized model experimented with different masking strategies. The results showed that the model performed best with a masking strategy of patch size of 64 and masked ratio of 0.5 after transfer learning optimization. Therefore, this paper uses this masking strategy to train the augmented dataset. The final model obtained is SL-DDBD. In this section, we will compare the accuracy of the baseline, the optimized model trained with original data (Improved), the model trained with the vision Transformer (ViT) as the encoder, and the Improved+DA model at different epochs, as shown in Fig. 10. The Improved+DA model has the fastest convergence speed and the second is the ViT model. However, the accuracy of the ViT model at the end is the lowest, only 74.35%. It is obvious that the Improved+DA model has reached a high accuracy of 78% at epoch 10. At the same epoch, the accuracy of other models is basically kept between 15%-25%. The accuracy of the Improved model is higher than the baseline at all epochs, which again proves the improvement of the optimization work on the model accuracy. The Improved+DA model not only converges quickly but also outperforms other models at each epoch. The accuracy of the model is 99.60%.

At the same time, in this experiment, the baseline model and our proposed model SL-DDBD are compared. The performance of the two models is compared in eight metrics: Acc, Ws, GFLOPS, Params, FPS, Precision, Recall, and F1-score. As shown in Table V, Fig. 11, Fig. 13, and Fig. 14. The confusion matrix of the SL-DDBD model is shown in Fig.12. After the optimization work, our model not only significantly improved the accuracy but also reduced the overall parameters and computation.

TABLE V
COMPREHENSIVE PERFORMANCE OF SL-DDBD AND BASELINE

| Model | Acc (%) | $W_s$/MB | GFLOPS | Parameters | V/FPS |
|---|---|---|---|---|---|
| Baseline | 84.92 | 994.7 | 15.26 | 86753474 | 29 |
| **SL-DDBD** | **99.60** | **316.3** | **4.49** | **27527044** | **56** |

The accuracy of the proposed model SL-DDBD reaches the highest level of 99.60%, which is an improvement of 14.68% compared to the baseline model's accuracy of 84.92%. SL-DDBD has excellent results in metrics of F1-score, Precision, and Recall. Especially for categories C8 and C9, the original baseline recognition accuracy was low. However, SL-DDBD significantly improved this situation, maintaining high values for Precision and F1-score across all categories. The FPS of the model has also improved after lightweight optimization, from the original 29 FPS to 56 FPS, which is nearly a two-fold increase. This not only ensures fast detection but also meets the real-time monitoring requirements. The detection performance of the model has reached an acceptable level in terms of both accuracy and detection speed.

common encoder structure, the model itself is not very lightweight in terms of parameters and weight file size. However, through lightweight optimization in the transfer learning section, the parameters of the model were reduced from 86.7 million to 27.5 million, a decrease of 68.2%. The size of the weight file after training was reduced from 994.7 MB to 316.3 MB. GFLOPS decreased from 15.26 to 4.49, which greatly reduces the difficulty in engineering deployment and hardware requirements.

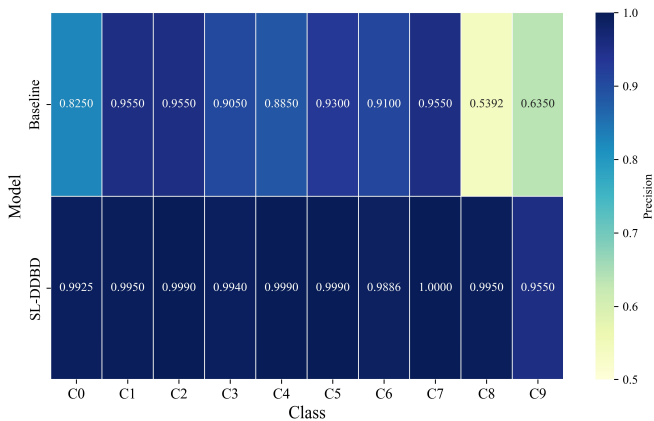

**Fig. 11.** Heat map comparison of precision for the baseline model and SL-DDBD.

Since the self-supervised learning framework was originally designed to solve large-scale computer vision tasks and used a

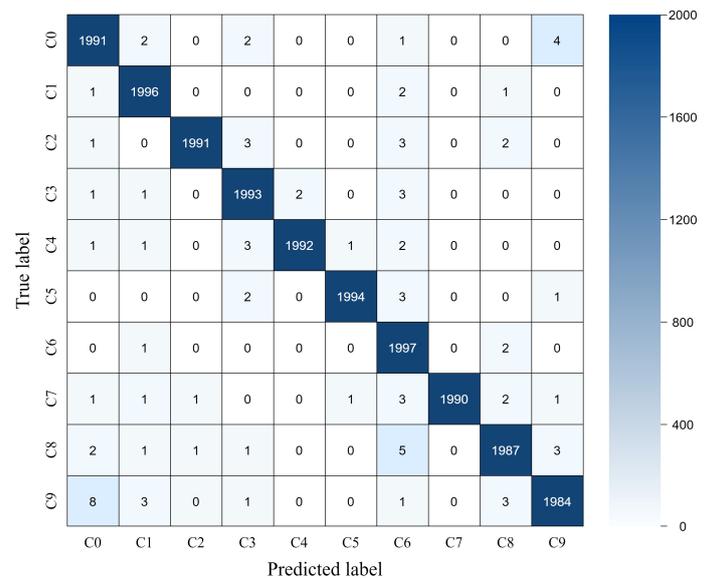

**Fig. 12.** Confusion matrix of SL-DDBD.

In the confusion matrix, the overall classification of the model is excellent, but it also shows a small number of misclassifications. For example, C0 "Normal driving" and C9

"Talking to passenger" are confused with each other. The main reason for the classification error is the high similarity of the images between the two categories. Drivers occasionally turn their heads during safe driving to observe the front view or to look at the rearview mirror. However, this may be misjudged as talking with passengers. Similarly, in some images of C9 "Talking to passenger", drivers do not obviously turn their heads to talk to passengers on the right, which is also easily misjudged as "Normal driving".

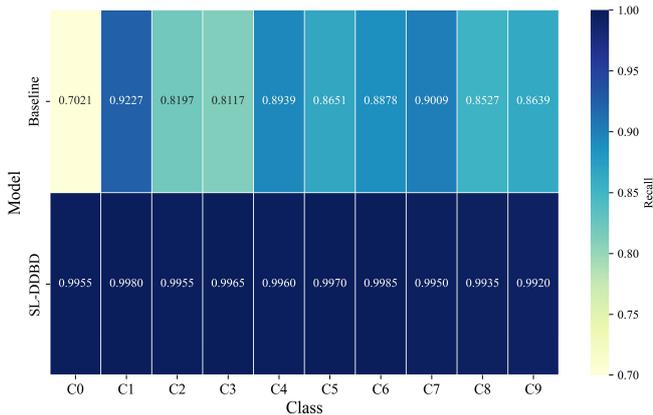

**Fig. 13.** Heat map comparison of recall for the baseline model and SL-DDBD.

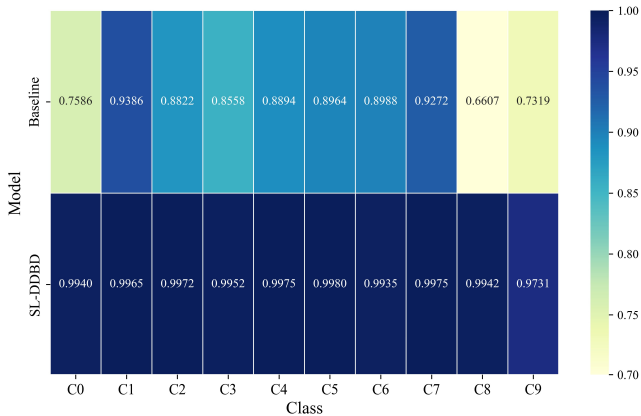

**Fig. 14.** Heat map comparison of F1-scores for the baseline model and SL-DDBD.

### E. Comparison with Previous Results

To validate the detection performance of the proposed method, we compared our best model, the "SL-DDBD" model, with previous state-of-the-art supervised and unsupervised learning methods.

The comparison results between the proposed method and supervised learning models are shown in Table VI. The main evaluation metrics are accuracy and parameters, which are used to compare and analyze the model's accuracy and size.

TABLE VI
COMPARISON RESULTS WITH SUPERVISED MODELS

| Model | Parameters(in Millions) | Acc(%) |
|---|---|---|
| Alexnet+Softmax[55] | 63.2 | 96.8 |
| Alexnet+Tripletloss[55] | 63.2 | 98.6 |
| D-HCNN[26] | 0.76 | 99.87 |
| HCF[56] | >72.3 | 96.74 |
| OLCMNet[57] | - | 89.53 |
| VGG-GAP[58] | 140 | 98.7 |
| Ensenmble VGG-16 and VGG-16[58] | >140 | 92.6 |
| Driver-net[59] | - | 95.0 |
| Vanilla CNN with data augmentation[60] | 26.05 | 97.05 |
| InceptionV3+Xception-50+Xception+VGG-19[61] | 214.3 | 97.00 |
| **SL-DDBD** | **27.5** | **99.60** |

"-" No data available

Literature [55] and [58] both use earlier CNN networks and improve the accuracy through some improvements. Their accuracy has not reached a high value. At the same time, due to the early network, the feature learning ability is poor, which leads to larger network parameters. Models like Ensemble VGG-16 and VGG-16 [58] have already exceeded 140 million in terms of parameters, but the accuracy is still a low value of 92.6%. The method in literature [61] is a fusion of multiple detection models, with parameters reaching 214.3 million, and the accuracy is still 97%. Similarly, the HCF [56] method, which is based on the fusion of multi-category CNN models, reduces the parameters to the lowest value of 72.3 million, but the accuracy is still not high. Recently, the D-HCNN [26] method has achieved high accuracy, exceeding 99%, by decreasing the convolutional kernel size to lightweight. Due to the large dataset of self-supervised training and the encoder of the Swin Transformer, the parameters of the basic model are larger compared to CNN models. However, the proposed method in our work is based on self-supervised learning, which is a particular advantage. The cost of supervised learning training will be significantly reduced, and it has strong generalization capabilities. Transfer learning only requires a small dataset and performs well in more detailed downstream tasks. Moreover, we have conducted lightweight work and verified its effectiveness through experiments. Finally, the model's parameters can be reduced to 27.5 million while achieving a recognition accuracy of 99.6%. Compared to most of the supervised learning-based CNN models in the table, the recognition accuracy is higher and the parameters are smaller.

Currently, there are only a few unsupervised learning methods available in the field of driver distraction behavior recognition. This indicates the need for further exploration and research in driver behavior detection. This paper compares the proposed method with recent unsupervised learning methods for driver distraction behavior detection [49], and presents the comparison results in Table VII.

TABLE VII
Comparison Results with Unsupervised Models

| Method | Backbone | Batch-size | Epoch | Acc(%) |
|---|---|---|---|---|
| Simsiam | ResNet50 | 32 | 400 | 86.29 |
| SimCLR | ResNet50 | 32 | 400 | 94.32 |
| UDL | RepMLP-Res50 | 32 | 400 | 98.61 |
| Baseline | Swin Transformer | 32 | 110 | 84.92 |
| Improved | Swin Transformer | 32 | 110 | 95.13 |
| **SL-DDBD** | **Swin Transformer** | **32** | **110** | **99.60** |

The SimCLR is a classic unsupervised learning method, with a ResNet50 backbone achieving 94.32% accuracy. The Simsiam method, also with a ResNet50 backbone, has an accuracy of 86.29%. Through the process of model refinement, our approach has surpassed the accuracy of both Simsiam and SimCLR, achieving a high recognition rate of 95.13%. The baseline model's recognition performance is only 1.37% lower than that of Simsiam. However, it's worth noting that, unlike the models presented in this work which were trained for 110 epochs, all other models underwent a significantly larger number of training epochs, reaching up to 400. In the SimCLR unsupervised learning method, a larger batch size provides more negative examples to facilitate convergence. Longer training epochs also provide more negative examples and can significantly improve results. All models in this work have only 110 training epochs. Masked image modeling is a new "prediction" type of self-supervised learning method. This method has good representation learning ability supported by pre-training, reducing the cost of transfer learning for driver distraction behavior detection. The UDL model proposed in the literature [49] is an improvement on the Simsiam structure and is also a "constructive" category unsupervised learning method like SimCLR. With the main network using RepMLP-Res50, the final accuracy of the UDL model is 98.61%. The accuracy of the SL-DDBD model in this paper is 99.60%, which is higher than the recognition accuracy of the all above models.

## V. Conclusion

The paper explores the introduction of MIM self-supervised learning method in the task of driver distraction behavior detection. Image masking strategy is used for pre-training on a large number of unlabeled datasets. In order to better integrate unsupervised learning and downstream tasks, transfer learning is carried out on the driver distraction behavior dataset. Lightweighting and accuracy optimization work has been done in transfer learning.

(1) The original encoder was improved through a reconfiguration of the number of Swin Transformer blocks in stage 3. The task detection accuracy was improved while reducing the complexity of the encoder network.

(2) For each stage of the encoder, a new distribution of the number of W-MSA and SW-MSA detection heads was made. The number of feature transfers was reduced and attention was increased on key feature information.

(3) In transfer learning, the impact of MIM strategy on downstream tasks was considered and a comprehensive comparison experiment was designed. The best masking strategy was selected.

(4) We used a multi-class data augmentation strategy to simulate real-world scenarios to expand dataset. This further improved the model's generalization ability in complex scenarios. SL-DDBD achieves 99.60% accuracy on the large-scale driver distraction behavior dataset State-Farm.

In future work, interesting work is to try using multi-source information fusion methods for driver distraction behavior detection. we will predict the driver's eye focus position. The predicted eye focus position and the action behavior recognition results are combined to achieve more accurate driver distraction recognition. In addition, we will consider using model pruning methods to further do lightweight research on the self-supervised driver distraction behavior detection model. The model quantifies the weights during transfer learning training, and then deletes low-weight parts according to certain standards. Deploying the model on mobile devices for fast and real-time detection of driver distraction.